\documentclass[nohyperref]{article}

\usepackage{microtype}
\usepackage{graphicx}
\usepackage{subfigure}
\usepackage{booktabs} 
\usepackage[inline]{enumitem}
\usepackage[T1]{fontenc}

\usepackage{hyperref}

\usepackage[accepted]{icml2023}

\usepackage{amsmath}
\usepackage{amssymb}
\usepackage{mathtools}
\usepackage{amsthm}
\usepackage{multirow}
\usepackage{textcomp}

\usepackage[capitalize,noabbrev]{cleveref}

\theoremstyle{plain}

\theoremstyle{definition}

\theoremstyle{remark}

\usepackage[textsize=tiny]{todonotes}
\Crefname{equation}{Eq.}{Eqs.}

\icmltitlerunning{Enforcing representational dissimilarity to learn new features}

\begin{document}

\twocolumn[
\icmltitle{Exploring new ways: Enforcing representational dissimilarity to learn new features and reduce error consistency}



\icmlsetsymbol{equal}{*}

\begin{icmlauthorlist}
\icmlauthor{Tassilo Wald}{hi,dkfz}
\icmlauthor{Constantin Ulrich}{dkfz}
\icmlauthor{Fabian Isensee}{hi,dkfz}
\icmlauthor{David Zimmerer}{hi,dkfz}
\icmlauthor{Gregor Koehler}{hi,dkfz}
\icmlauthor{Michael Baumgartner}{hi,dkfz,hi}
\icmlauthor{Klaus H. Maier-Hein}{hi,dkfz,unihd}
\end{icmlauthorlist}

\icmlaffiliation{hi}{Helmholtz Imaging}
\icmlaffiliation{dkfz}{Department of medical image Computing, German cancer research center (DKFZ), Heidelberg, Germany}
\icmlaffiliation{unihd}{Pattern Analysis and Learning Group, Department of Radiation Oncology, Heidelberg University Hospital, Heidelberg, Germany}

\icmlcorrespondingauthor{Tassilo Wald}{tassilo.wald@dkfz-heidelberg.de}

\icmlkeywords{Representations, Similarity, Diversity}

\vskip 0.3in
]



\printAffiliationsAndNotice{} 

\begin{abstract}
Independently trained machine learning models tend to learn similar features. Given an ensemble of independently trained models, this results in correlated predictions and common failure modes. Previous attempts focusing on decorrelation of output predictions or logits yielded mixed results, particularly due to their reduction in model accuracy caused by conflicting optimization objectives. In this paper, we propose the novel idea of utilizing methods of the representational similarity field to promote dissimilarity during training instead of measuring similarity of trained models. To this end, we promote intermediate representations to be dissimilar at different depths between architectures, with the goal of learning robust ensembles with disjoint failure modes. We show that highly dissimilar intermediate representations result in less correlated output predictions and slightly lower error consistency, resulting in higher ensemble accuracy. With this, we shine first light on the connection between intermediate representations and their impact on the output predictions.
\end{abstract}

\section{Introduction}
\label{sec:intro}
Machine learning methods learn features automatically when trained on a dataset.
The high dimensionality of the data should allow for a variety of solutions, yet independent models tend to learn similar features to each other. Many downstream effects of this feature similarity are observed throughout current machine learning literature:
\begin{enumerate*}[label=\alph*)]
\item \citet{Geirhos2020a} showed that independently trained CNNs tend to predict erroneously on the same cases much more often than expected by chance given their accuracy, and more often than e.g. humans. 
\item \citet{Csiszarik2021, Bansal2021} showed that different models are functionally similar, through \textit{stitching} \cite{Lenc2015} the top of a model to the bottom of another independently trained model with marginal accuracy penalties. 
\item \citet{Ainsworth2022} showed that independently trained ResNets exhibit a linear mode connectivity with zero loss barrier, given a previous functionally invariant kernel weight permutation.
\item \citet{Moschella2023} showed that distinct latent spaces of two independently trained models tend to differ just by an quasi-isometric transformation.
\end{enumerate*}

\begin{figure}
    \centering
    \includegraphics[width=.7\linewidth]{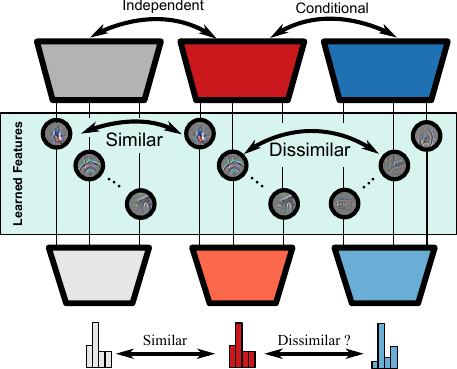}
    \caption{Two independently trained models with different random seed learn very similar features. We propose to train a novel model conditioned on an already trained model with an auxiliary loss enforcing dissimilarity at an intermediate processing stage. 
    }
    \label{fig:overview_fig}
\end{figure}

While the feature similarity is not a problem for a single model, multiple models are often combined into an ensemble to improve performance and to measure predictive uncertainty \cite{Lakshminarayanan2016}. When these models learn the same features, they may learn spurious correlations that are not actually useful for the task at hand. This causes them to share failure modes making them fail in the same way.
Ensemble improvement is highly dependent on models having a large disagreement error ratio \cite{Theisen2023} or low error consistency \cite{Geirhos2020a}. This can be increased slightly through different augmentation schemes, moderately through different pre-training schemes and strongly through pre-training on a different dataset, with higher error inconsistency in error rates improving ensemble benefits more \cite{gontijo-lopes2022no}.


Trying to find a new method to increase such predictive diversity between existing models and a new model may become difficult when learning large groups of models, hence methods were introduced that are conditioned on pre-existing models with the intention of learning to solve the task differently.

Early works explored negative correlation  \cite{Liu1999a,Liu1999b,Islam2003} and evaluated classical methods like boosting or bagging \cite{Dietterich2000} for neural networks under the assumption of them being weak classifiers.
These approaches suffer great accuracy penalties in modern settings, hence other approaches needed to be developed.
    
In a more modern setting \citet{Pang2019} trained multiple models simultaneously, enforcing high entropy and orthogonality between the negative class predictions leading to improved adversarial robustness and in-distribution performance of ensembles.

\citet{Minderer2020} proposed a two stage approach for the domain of adversarial learning, first training a model, then learning an augmenting adversarial auto-encoder, that tries to remove the most predictive features while staying as close as possible to the actual image. Given this lens a model can be trained with the lens to learn different features.


So far diversification approaches only regularize inputs or at the position of the output features. However the constraints on regularization at the input and the output are rather large. Adapting input images too much can degrade performance too much and the features at the very end are constrained by having to encode the target classes, constraining the potential solution space models can converge to.

In this paper, we propose to regularize internal representations of a new model to be dissimilar to an existing model to promote discovering novel ways of solving the task, which, to the best of our knowledge, has not been explored so far. Through this we hope to learn about the connection of internal similarity to the predictive behavior between models, specifically whether inducing diversity in intermediate processing stages leads to different predictive behavior and more robust ensembles.

Our main contributions are:
\begin{enumerate}
    \item We utilize methods from the field of representational similarity in a novel way to train ensembles of very low representational similarity at intermediate layers.
    \item We show that highly dissimilar internal representations can be learned at chosen positions with only minor penalties to the model accuracy.
    \item We show that enforcing dissimilar internal representations can lead to lower error consistency in the predicted outputs, overall improving ensembling performance relative to an ensemble of independently trained models.

\end{enumerate}

\section{Representational (dis)similarity}
\label{sec:representational_similarity}
 We are interested in enforcing models to learn different features termed representations, therefore we must define what a representation is.
 In the field of representational similarity representations $\textbf{z}$ are defined as the input-output responses of all channels of a network's layer to a set of inputs \cite{Li2016, Raghu2017, Morcos2018, Wang2018, Kornblith2019}.
 For a given input batch $x_i$ of $B$ samples this results in $\textbf{z} \in \mathcal{R}^{B\times C \times W \times H}$ values for each model.

 \begin{figure}
    \centering
    \includegraphics[width=.9\linewidth]{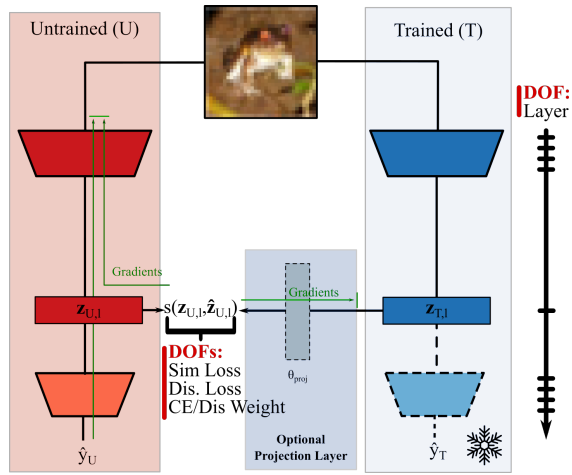}
    \caption{We enforce representational dissimilarity during training of an untrained model (U) given an already trained model (T) through penalizing similarity for U (\cref{eq:full_loss_untrained}) and penalizing dissimilarity for T (\cref{eq:full_loss_trained}). As depicted there exist many Degrees of Freedom (DOF) on how to design the system, of which we explore a variety.
    }
    \label{fig:training_scheme}
\end{figure}

 Since e.g. the channel dimensions between neurons are not aligned \cite{Li2016} metrics either need to learn a linear transformation to align representations or use sub-space metrics, e.g. canonical correlation analysis \cite{Raghu2017, Morcos2018}, to measure similarity. For a nice summary of the differences we refer to \citet{Kornblith2019} and more recently \citet{Klabunde2023}. 

To formalize: An already trained model $T$ and an untrained model $U$ are composed of $k$ sequential layers $f_{\theta}$ predicting some class probabilities $\hat{y}$, see \cref{eq:composition}.
\begin{equation}
    \hat{y}_{U,i} = (f_{\theta_{U,k}}\circ \cdots \circ f_{\theta_{U,0}})(x_i)\\
    \label{eq:composition}
\end{equation}
At some intermediate layer of choice $l$ we collect representations of both models $\textbf{z}_{T,i,l}$ and $\textbf{z}_{U,i,l}$,
\begin{equation}
    \textbf{z}_{U,i,l} = (f_{\theta_{U,l}}\circ \cdots \circ f_{\theta_{U,0}})(x_i)
    \label{eq:first_split_composition}
\end{equation}
\begin{equation}
    \hat{y}_{U,i} = (f_{\theta_{U,k}} \circ \cdots \circ f_{\theta_{U,l-1}})(\textbf{z}_{U,i,l})
    \label{eq:last_split_composition}
\end{equation}
and measure similarity $s$ between $\textbf{z}_{T,i,l}$ and $\textbf{z}_{U,i,l}$, using a similarity metric $\mathcal{S}$ providing a bounded value between $a$ and $b$ with the lowest value, $a$, representing most dissimilar and $b$ most similar.
$\mathcal{S}(\textbf{z}_{T,l}, \textbf{z}_{U,l})~ \text{with }~\mathcal{S}(\textbf{z}_1,\textbf{z}_2) \in (a, b) ~\text{for }\forall ~\textbf{z}_1,\textbf{z}_2$. \footnote{We omit the batch index for better readability where appropriate.}

Similarity metrics like channel-wise correlation and regression need aligned channel pairs between the two representations, hence we introduce a projection layer $proj_{\theta_{p}}(\textbf{z}_T)$ that tries to approximate the representations of the untrained $\textbf{z}_U$ by linearly combining the activations $\textbf{z}_T$ of the same spatial location, through a linear projection resulting in 
\begin{equation}
\hat{\textbf{z}}_{U,l} = p_{\text{proj}}(\textbf{z}_{T,l}).
\label{eq:approximation}
\end{equation}

Combining these, the final objectives of the new model $U$ can be described as
\begin{equation}
\begin{aligned}
    \min_{\theta_U}\mathcal{J}(y_i, x_i,\textbf{z}_{T,i,l}) &=  ~\text{CE}(\hat{y}_i, y_i) \\
    &+ \lambda ~ \mathcal{S}(\textbf{z}_{U,i,l}, \hat{\textbf{z}}_{U,l}),
    \label{eq:full_loss_untrained}
\end{aligned}
\end{equation}
with lambda being a weighting factor for controlling the importance of learning dissimilar representations. Simultaneously the projection layer is optimized to maximize similarity between the representations of $T$ and $U$. 
\begin{equation}
    \min_{\theta_{proj}}\mathcal{J}(y_i, x_i, \textbf{z}_{U,i}) = - \mathcal{S}\left(\textbf{z}_{U,i}, \hat{\textbf{z}}_{U,l}\right).
    \label{eq:full_loss_trained}
\end{equation} A visualization of the training scheme is highlighted in \Cref{fig:training_scheme}.
To extend this scheme to an arbitrary number of models we concatenate the representations of an arbitrary amount of pre-existing models of a layer $\textbf{Z}_{T,l} \in \left\{\textbf{z}_{T_1,l}, \cdots, \textbf{z}_{T_N,l}\right\}$ before the projection layer. When training multiple models we train sequentially, ending up with one unregularized model, the first one, and multiple regularized models conditioned on all preceding models of the same sequence.

As similarity metrics $\mathcal{S}$ we evaluate the L2 Correlation \textit{L2Corr}, bounded explained variance of a linear regression \textit{ExpVar} and Linear Centered Kernel Alignment \textit{LinCKA} \cite{Kornblith2019}. The former two representing aligned metrics with \textit{L2Corr} being scaling invariant and the Explained variance being scaling sensitive. Full details of the functions are provided in \Cref{apx:metrics}.

\section{Experiments}
We train sequences of up to $5$ ResNets on CIFAR10, CIFAR100 and compare similarity and predictive behavior of the models within a sequence to each other and between models of different sequences. Explicit hyperparameters of the experiments are provided in \cref{apx:training_details}. Final similarity measurement between models is conducted through \textit{LinCKA} due to it not requiring a linear projection to be calculated. In the main paper we focus only on CIFAR10 and \textit{ExpVar} with $\lambda = 1$ unless stated otherwise, while the additional architectures and metrics can be found in \cref{apx:dissimilarity_ablations}.

\subsection{Internal dissimilarity can be controlled precisely}
\label{sec:dissimilarity_ablation}
Given the scheme proposed in \cref{fig:training_scheme} dissimilarity can be enforced in a variety of ways given the degrees of freedom, like layer position, dissimilarity weight $\lambda$, different regularization metrics, regularizing multiple layers simultaneously, effects of larger models or other datasets. Of these experiments we only highlight the Layer position here while the remaining ones can be found in \cref{apx:dissimilarity_ablations}.

\begin{figure}
    \centering
    \includegraphics[width=.7\linewidth]{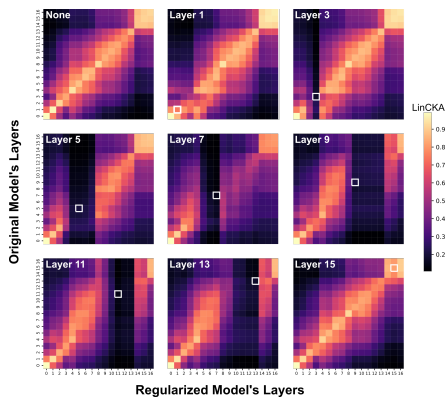}
    \caption{We compare ResNet34s $T$ and $U$ through \textit{LinCKA} after learning dissimilar representations through \textit{ExpVar} at a layer of choice. Enforcing representational dissimilarity at different layers leads to highly dissimilar representations between models. Early and later layers remain more similar while intermediate layers can easily be regularized. A visualization of the diagonal \textit{LinCKA} values can be found in \cref{fig:ablation_position_in_architecture_diagonal}.}
    \label{fig:diff_layers_modded_v2}
\end{figure}

\paragraph{Layer position}
Many different layers can be selected to enforce representational dissimilarity, therefore we evaluate a variety of positions (see \cref{fig:diff_layers_modded_v2}). We find that regularizing earlier and later layers leads to a lower decrease in similarity, while intermediate layers easily become almost fully dissimilar. 
Interestingly we observe that the effect of the regularization often translates to neighboring layers being also highly dissimilar, mostly affecting the entire residual block, becoming more similar again after a down-sampling layer.


\subsection{Dissimilar representations are unique}
\begin{figure}
    \centering
    \includegraphics[width=.9\linewidth]{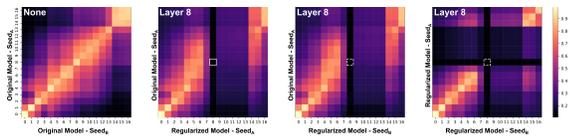}
    \caption{From left to right \textit{LinCKA} of ResNet34 models regularized with \textit{ExpVar}:
        1) Unregularized models compared to each other.
        2) An unregularized model compared to a regularized model from its sequence.
        3) An unregularized model compared to a regularized model from a different sequence.
        4) A regularized model to another regularized model from a different sequences.
        Since models from different sequences are not enforced to be dissimilar we denote the layer of regularization as dashed .} 
        \label{fig:unique_regularizations}
\end{figure}

Given that independent models converge to very similar solutions, one might assume that models, trained to be dissimilar from them may end up being highly similar to each other. This assumption seems to be amiss, as we observe high representational dissimilarity between the models, see \cref{fig:unique_regularizations}. The regularized models are even more dissimilar to each other than they are to the base model they were regularized on, highlighting that the different models learn unique dissimilar solutions to the baseline models.

\subsection{Ensembles of dissimilar models}
\begin{figure}
    \centering
    \includegraphics[width=.5\linewidth]{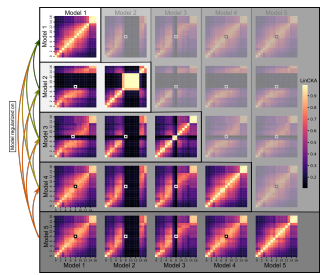}
    \caption{LinCKA between all members of a single ensemble regularized with ExpVar at layer 8. The models are trained sequentially with every new model $U$ being regularized on all existing models.} 
        \label{fig:examplary_group_of_5_ensembles}
\end{figure}

After establishing that models can learn dissimilar representations we extend the setting to larger number of models training an ensemble of  $N=5$ ResNet34s and compare them to each other as visualized in \cref{fig:examplary_group_of_5_ensembles}. We observe high dissimilarity between early models at the chosen layer, while later models have a less pronounced similarity decrease to existing models. 
We attribute this to the fact that all representations are merged to approximate the new model, resulting in the new models learning dissimilar representations to this group and not every single model. 
While this effect looks like efficacy decreases with growing numbers, the model still learns lower similarity levels than the independently trained baseline.

\subsection{Downstream effects of internal dissimilarity}
Due to the compositional nature of networks (\cref{eq:composition}) we wonder if high internal dissimilarity translates to less correlated predictive behavior between the model pairs. To this end we compare the absolute ensemble performance, the error consistency Cohen's Kappa $\kappa$ between models (see \cref{apx:output_metrics}) and the accuracy of the latest trained model $Acc_U$ for an ensemble of up to 5 models, see \cref{tab:ensemble_performance_main}.
We observe that regularizing representational similarity leads to an overall decrease in Cohen's Kappa of $3\%$ up to $7\%$, while largely maintaining single model performance. Subsequently the ensemble composed of dissimilar models features higher ensemble performance than the baseline ensemble of independently trained models, for every number of models in the ensemble. 

\begin{table}
\centering
\caption{Ensembles of ResNet34, regularized at various layers with the \textit{ExpVar} metric. Best metrics are displayed bold, while metrics better than the baseline are underlined.}
\resizebox{\linewidth}{!}{
\begin{tabular}{ccccccc}
\toprule
 \multirow{2}{*}{\textbf{Output metric}} &  \multicolumn{2}{c}{\textbf{Dissimilarity}} & \multicolumn{4}{c}{\textbf{Ensemble of N Models}}\\
 \cmidrule{4-7}
 & \textbf{Metric} & \textbf{Layer} & 2 & 3 & 4 & 5 \\
\midrule
\multirow{5}{*}{\textbf{Ensemble Acc.} [$\%$] $\uparrow$} & \textbf{Baseline} & - & 95.438 \textpm 0.087 & 95.730 \textpm 0.082 & 95.880 \textpm 0.073 & 95.936 \textpm 0.081 \\
& \multirow{4}{*}{\textbf{ExpVar}} & 1 & \underline{95.580} \textpm 0.189 & \textbf{95.914} \textpm 0.039 & \underline{96.014} \textpm 0.048 & \textbf{96.130} \textpm 0.042 \\
&  & 3 & \underline{95.543} \textpm 0.147 & \underline{95.877} \textpm 0.118 & \underline{95.968} \textpm 0.112 & \underline{95.975} \textpm 0.038 \\
&  & 8 & \underline{95.570} \textpm 0.010 & \underline{95.897} \textpm 0.006 & \textbf{96.050} \textpm 0.092 & \underline{96.103} \textpm 0.025 \\
&  & 13 & \textbf{95.748} \textpm 0.114 & \underline{95.906} \textpm 0.113 & \underline{95.970} \textpm 0.029 & \underline{96.034} \textpm 0.076 \\
\midrule
\multirow{5}{*}{\textbf{$Acc_{U}$} [\%] $\uparrow$} & \textbf{Baseline} & - & \textbf{94.708} \textpm 0.223 &  94.828 \textpm 0.165 & 94.918 \textpm 0.163 & \textbf{94.898} \textpm 0.134 \\
& \multirow{4}{*}{\textbf{ExpVar}} & 1 & 94.426 \textpm 0.292 & 94.810 \textpm 0.079 & \textbf{94.942} \textpm 0.100 & \textbf{94.896} \textpm 0.302 \\
& & 3 & 94.325 \textpm 0.254 & \textbf{94.903} \textpm 0.043 & 94.875 \textpm 0.186 & 94.832 \textpm 0.163 \\
& & 8 & 94.403 \textpm 0.146 & 94.733 \textpm 0.377 & 94.707 \textpm 0.095 & 94.770 \textpm 0.425 \\
& & 13 & 94.608 \textpm 0.304 &  94.772 \textpm 0.173 & 94.828 \textpm 0.278 & 94.636 \textpm 0.204  \\

\midrule
\multirow{5}{*}{$\kappa$ [$\%$] $\downarrow$} & \textbf{Baseline} & - & 52.045 \textpm 0.707 & 51.852 \textpm 1.417 & 51.478 \textpm 1.226 & 51.588 \textpm 1.113 \\
& \multirow{4}{*}{\textbf{ExpVar}} & 1 & \underline{45.636} \textpm 1.696 & \underline{47.153} \textpm 1.162 & \underline{48.052} \textpm 0.683 & \underline{48.377} \textpm 1.035 \\
&  & 3  & \underline{45.472} \textpm 1.844 & \underline{46.818} \textpm 0.931 & \underline{48.125} \textpm 0.498 & \underline{48.601} \textpm 0.542 \\
&  & 8  & \underline{45.358} \textpm 0.839 & \textbf{46.438} \textpm \underline{0.801} & \textbf{46.838} \textpm 1.217 & \textbf{46.776} \textpm 1.590 \\
&  & 13  & \textbf{44.361} \textpm 0.803 & \underline{46.771} \textpm 0.708 & \underline{48.169} \textpm 0.772 & \underline{48.692} \textpm 0.327 \\
\bottomrule
\end{tabular}%
}
\label{tab:ensemble_performance_main}
\end{table}

\section{Discussion, Limitations and Conclusion}
In this paper we enforce models to learn very dissimilar internal representations to preexisting models by applying various metrics from the field of representational similarity in a novel way. We show that learning features that can't be approximated through linear regression at an intermediate layer position can result in lower error consistency between the two models. Furthermore we show that this slight decrease in error-consistency suffices to improve overall ensemble accuracy over an ensemble of independent models.

While we show that error consistency between models can be decreased through representational dissimilarity, the decrease is still slight, showing a decrease between 7\% to 3\% points. Optimally one would like to achieve Cohen's Kappa scores of $<0$ to maximize ensembling benefits. 

Furthermore, we could explore only a small subset of all experiments we deem interesting and have to leave a lot of interesting questions for future work, e.g.: \begin{enumerate*}[label=(\alph*)]
    \item Which features does the dissimilar model learn?
    \item Does the dissimilar model end up in a disjoint loss minimum? Can it still be stitched?  
    \item Which position reduces Cohen's Kappa the most while maintaining single model performance?
    \item Are there other metrics that decrease error consistency more strongly with less effect on accuracy?
\end{enumerate*}

Overall, we provide a proof of concept, that enforcing models to learn dissimilar representations also results in different predictive behavior, providing another avenue to make models learn different, novel  features, with the goal of achieving robust ensembles. 

\bibliography{example_paper}
\bibliographystyle{icml2023}
 
\newpage
    
\newpage
 
\appendix
\section{Experiment Details}
\label{apx:training_details}
In this paper the architectures we use are ResNet18s, ResNet34s and ResNet101s trained on CIFAR10 and CIFAR100. 
Across all experiment settings we train 5 indepedent runs that differ by their random seed, including the ensemble sequence experiments. 

\subsection{Hyperparameters}
All hyperparameters were optimized to maximize single model validation accuracy, as one would in a normal training setting. Finding best performing hyperparameters is a crucial step in order to assure that the developed methodology does not only work for less accurate models where more predictions errors are made and the diversification task becomes easier. We conducted this process for all datasets separately, with all architectures sharing the same final hyperparameter settings.

Once found on single models the hyperparameters were frozen and not adapted to the dissimilarity regularization scheme, avoiding optimizing hyperparameters when starting the dissimilarity regularization experiments. The following hyperparameters for the different architectures were used:

\paragraph{CIFAR10} We train the architectures for $250$ epochs with a batch size of $128$, trained with nesterov SGD, learning rate of $0.1$, momentum $0.9$, cosine annealing learning rate schedule and weight decay $5e-4$. For augmentations we use RandomCrops of size $32\times32$ with padding $4$, the AutoAugment CIFAR10 policy from \citet{Cubuk2019}, followed by Cutout of with size $16$, finished by normalizing the images.
\paragraph{CIFAR100} We utilize the same hyperparameters from CIFAR10 just with a shorter epoch time of $200$.

\subsection{Architectures}
\begin{figure}
    \centering
    \includegraphics[width=.5\linewidth]{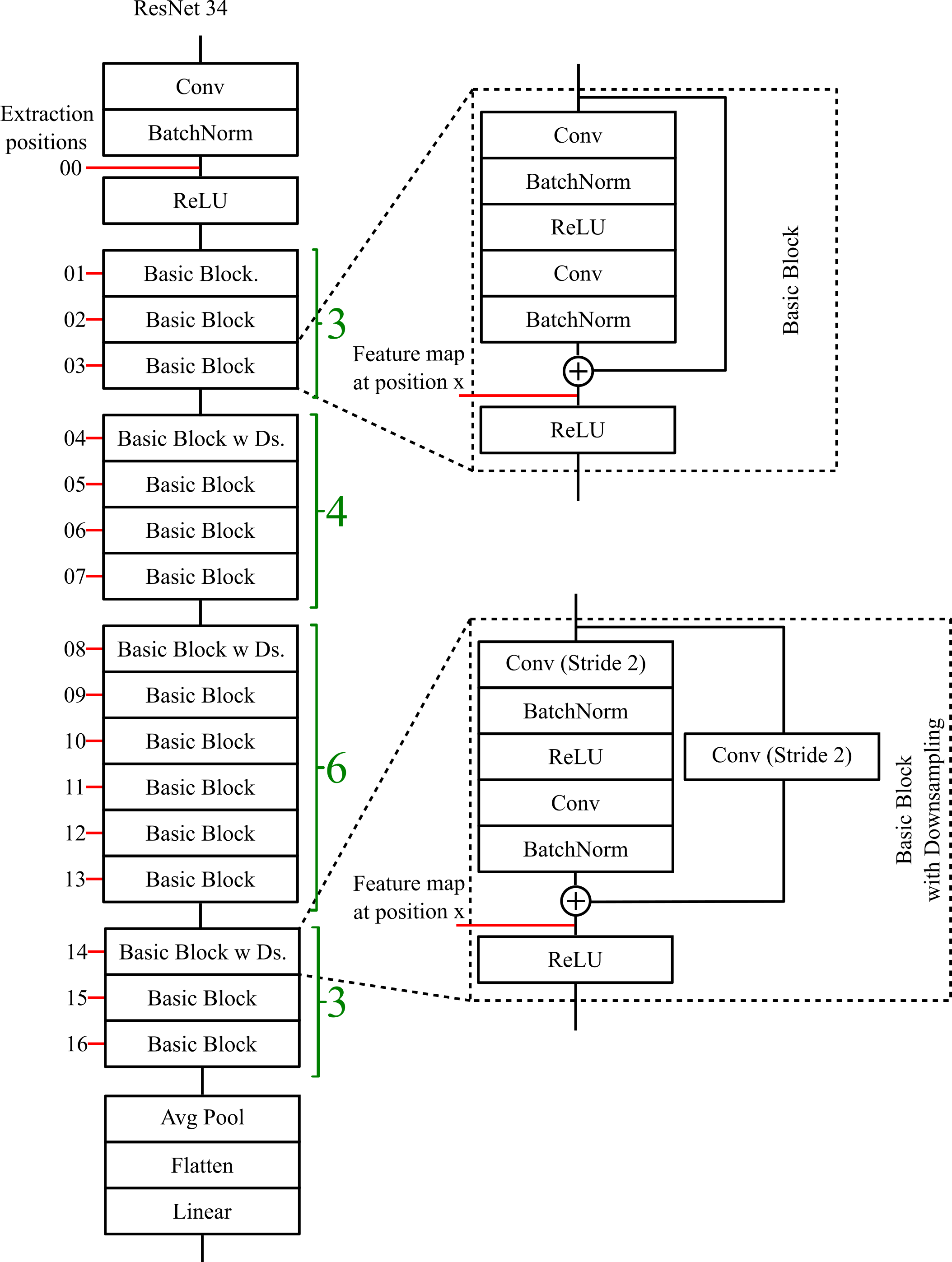}
    \caption{We extract the representations $\textbf{z}$ before the ReLU activation funciton at a location where no skip connection  circumvents the current layer. Exemplary ResNet34, using "basic blocks" with or without spatial reduction. We refer to this as ResNet34 with residual blocks of depth (3-4-23-3).}
    \label{fig:resnet_extraction}
\end{figure}

We train ResNet18, ResNet34 and ResNet101\footnote{The architectures implementation is inspired by \hyperlink{https://github.com/kuangliu/pytorch-cifar}{kuangliu's Github repository}}, with the extraction positions being before the $ReLU$ activation function, as highlighted in \cref{fig:resnet_extraction}. All ResNets \citep{He2016} are composed of 4 blocks containing various number of layers per block with a constant amount of channels per block. 

\paragraph{ResNet18}
Our ResNet18 is composed of four residual blocks of (2-2-2-2) layers with (64-128-256-512) channels each, using basic residual blocks without the inverse bottleneck structure. 

\paragraph{ResNet34}
ResNet34 is composed of four residual blocks of (3-4-6-3) layers with (64-128-256-512) channels each, using basic residual blocks without inverse bottleneck structure.

\paragraph{ResNet101}
ResNet101 is composed of four residual blocks of (3-4-23-3) layers with (256-512-1024-2048) channels respectively, using inverse bottleneck structure of blocks.

\subsection{Evaluation}
All results provided in this document are calculated on the test sets of the respective datasets. For CIFAR10 and CIFAR100 we use the official test set of $N=10000$ samples to evaluate both, representational similarity between models and output metrics between models.

For an ensemble $>2$, most of the output metrics like Cohen's Kappa $\kappa$, $JSD$ and $ERD$ are not directly defined. Hence, we calculate the pairwise metrics and average them over all pairs, leading to an average value which we report in e.g. \cref{tab:ensemble_performance_main}.

\section{Metrics}
\label{apx:metrics}
The metrics we employ in this study are threefold:
L2-Correlation (\textit{L2Corr}), explained variance (\textit{ExpVar}) and the sub-space metric Linear Centered Kernel Alignment (\textit{LinCKA}) \cite{Kornblith2019} as similarity measures $\mathcal{S}$.

\subsection{(Dis)similarity metrics}
\paragraph{Aligned metrics}
Some metrics compare channels directly and subsequently need exact alignment between the channels of the two networks $U$ and $T$. This channel alignment is non-existent when trained independently and needs to be learned. Hence, we utilize the linear projection to learn channel-wise alignment, as explained in \cref{sec:representational_similarity}. We implement this linear projection through a $1\times1$ Convolution, which maintains spatial information. 
Given these aligned, approximated values \cref{eq:approximation} of the new representation $\textbf{z}_U$, we can calculate the \textit{L2Corr} $r_c$ and \textit{ExpVar} $R^2$ for each channel.

\begin{equation}
    r_c(\textbf{z}_c, \hat{\textbf{z}}_c)={\frac {\sum _{i=1}^{n}(\textbf{z}_{i}-{\bar {\textbf{z}}})(\hat{\textbf{z}}_{i}-{\hat {\bar{\textbf{z}}}})}{{\sqrt {\sum _{i=1}^{n}(\textbf{z}_{i}-{\bar {\textbf{z}}})^{2}}}{\sqrt {\sum _{i=1}^{n}(\hat{\textbf{z}}_{i}-{\hat {\bar{\textbf{z}}}})^{2}}}}}
    \label{eq:l2corr}
\end{equation}

\begin{equation}
R^{2}_c(\textbf{z}_{c}, \hat{\textbf{z}}_c) = 1 - \frac{\sum_{i=1}^{n} (\mathbf{z}_{i} - \mathbf{\hat{\textbf{z}}}_{i})^2}{\sum_{i=1}^{n}(\mathbf{z}_{i} - \mathbf{\bar{z}})^2} 
\label{eq:exp_var}
\end{equation}

While the correlation is nicely bounded between $r_c \in \{-1, 1\}$, the explained variance $R^2$ is bounded upwards to 1 but can reach $-\inf$. Therefore we introduce a celu function that wraps the \cref{eq:exp_var} with $\alpha = 1$ that bounds the function.
\begin{equation}
    \begin{aligned}
        \text{CELU}(R^2)=&\max(0,R^2)\\
        +&\min(0,\alpha \cdot(\exp(R^2 / \alpha) - 1))
    \end{aligned}
    \label{eq:celu}
\end{equation}
In the conducted experiments this was necessary to assure stability of the convergence process, as single representation channels $\textbf{z}_{U,c}$ could occasionally feature very low variance close to $0$ in a mini-batch, leading to numeric problems when calculating $R^2$ scores. 

Given the channel wise scores we average them over all channels resulting in the final score.
\begin{equation}
    \text{\textit{ExpVar}} \coloneqq \frac{1}{C} \cdot \sum^C_{c=0} \text{CELU}(R^2_c)
\end{equation}
\begin{equation}
    \text{L2Corr} \coloneqq \frac{1}{C} \cdot \sum^C_{c=0} \text{CELU}(r_c)
\end{equation}

\paragraph{Sub-space metrics}
Layers of CNNs are composed of multiple channels with the directly following convolution combining the values of its preceding layer through a linear weighted sum of $k\times k\times C$ spanning all channels, making channel order irrelevant and a subject to the stochastic optimization process. 
This lack of alignment of single neurons was highlighted by \cite{Li2016} showing that no perfect one-to-one matching of single channels/neurons exists. This insight led to the field of sub-space metrics, which don't try to measure similarity between single channels/neurons but view the stack of channels as vectors spanning a sub-space \cite{Raghu2017, Morcos2018, Wang2018, Kornblith2019}. These spanned sub-spaces can then be compared directly. 

In this paper we use linear Centered Kernel Alignment (CKA) \citep{Kornblith2019} a metric inspired by the dissimilarity matrices of the neuroscience domain\citep{Kriegeskorte2008}, which does not need this channel alignment either.
Linear Centered Kernel Alignment itself leverages the Hilbert-Schmidt Independence Criterion (HSIC) to compare the similarity of such similarity matrices. In this work we use the unbiased HSIC estimator (\cref{eq:hsic}) introduced by \cite{Song2012} and used in mini-batch CKA (\cref{eq:cka}) as introduced by \cite{Nguyen2021}.

\begin{equation}
\begin{split}
    &\text{CKA}_{minibatch}(\textbf{K}, \textbf{L}) = \\
    &\frac{\frac{1}{k}{\sum_{i=1}^k} \text{HSIC}(K_i,L_i)}{\sqrt{\frac{1}{k}\sum_{i=1}^{k}\text{HSIC}(K_i,K_i)} \sqrt{\frac{1}{k}\sum_{i=1}^{k}\text{HSIC}(L_i,L_i)}}
    \label{eq:cka}
\end{split}
\end{equation}

\begin{equation}
\begin{split}
    &\text{HSIC}(\textbf{K},\textbf{L}) = \frac{1}{n(n-3)} \\
    &\cdot \left( 
    tr(\Tilde{\textbf{K}}\Tilde{\textbf{L}})
    + \frac{\textbf{1}^T \Tilde{\textbf{K}} \textbf{1}\textbf{1}^T \Tilde{\textbf{L}} \textbf{1}}{(n-1)(n-2)}
    - \frac{2}{n-2} \textbf{1}^T\Tilde{\textbf{K}}\Tilde{\textbf{L}}\textbf{1}
    \right)
    \label{eq:hsic}
\end{split}
\end{equation}

\subsection{Output prediction metrics}
\label{apx:output_metrics}

\paragraph{Cohen's Kappa}
Cohen's Kappa \cite{Cohen1960} measures the error consistency between predictors. Specifically it measures the observed error overlap $e_{obs}$ over a cohort $N$ and relates it to the expected error overlap $e_{exp}$ given the accuracies of the predictors while assuming independence between the raters. A $\kappa$ of 0 indicates that the observed error overlap between the two raters is exactly the expected value of error overlap given the two raters accuracies, indicating that the raters are independent. Values of 1 indicate that when one raters errs the other errs as well, whereas negative values indicate that if one rater errs the other is more likely to be correct on the case.
\begin{align}
\kappa &= \frac{e_{obs} - e_{exp}}{1 - e_{exp}}
\label{eq:CohensKappa}
\end{align}
Given an ensemble of models a low or negative value of $\kappa$ is highly desirable as it indicate that the models do not fail on the same samples, leading to high uncertainty on disagreed upon cases or given enough models a correct consensus.

Following up on this error-inconsistency was introduced, which measures when one or the other rater fails but not both. The error-inconsistency is a desirable feature as high values indicate that mistakes are disjoint and, given enough models in an ensemble, can be corrected to improve ensemble performance significantly \cite{gontijo-lopes2022no}.

\paragraph{Jensen-Shannon Divergence}
The Jensen-Shannon divergence (JSD) measures the similarity between two distributions and is based on the Kullbach-Leibeler-Divergence (KL-Divergence) $D$. Opposed to Cohen's Kappas, which only cares about the argmax prediction being right or wrong, it works with the probability distributions of two models instead, measuring changes in predictive behaviour in a less discrete way.
\begin{align}
    JSD(P\Vert Q) &= \frac{1}{2} \left( D(P\Vert M) + D(Q\Vert M)\right)\text{, with}\\
    M &= \frac{1}{2} (P + Q)
    \label{eq:JSD}
\end{align}
In our case, the probability distribution is represented by the output softmax probabilities of the models, hence the JSD measures how similar the distributions of the two models predictions are before assigning a hard label.

\paragraph{Error disagreement ratio}
Additionally to the established metrics, we propose a metric we term \textit{error disagreement ratio} (EDR), which measures the ratio of different errors to identical errors between two predictors for all joint errors  $N_{wrong}$ (\cref{eq:error_disagreement_ratio}). If two models disagree in their errors as often as they agree on their errors EDR is 1. Should the models always agree on their errors the EDR would be 0 and the EDR is greater than 1 for cases where the models disagree more often than agree when both err. 

Given that silent failures are detrimental for the applicability of deep learning methods, a high EDR is wanted, yet the EDR in our baselines is commonly $<1$ around $0.3$.

\begin{equation}
    \text{EDR}(\hat{p}_1, \hat{p}_2) = \frac{\sum^{N_{wrong}}_i (1 - eq(\hat{p}_{1_i}, \hat{p}_{2_i})}{\sum^{N_{wrong}}_i eq(\hat{p}_{1_i}, \hat{p}_{2_i})}
    \label{eq:error_disagreement_ratio}
\end{equation}

\begin{equation}
\text{eq}(\hat{p}_{1_i}, \hat{p}_{2_i}) =  \begin{cases}
                            1, & \text{if } \hat{p}_{1_i} = \hat{p}_{2_i} \\
                0 ,              & \text{else}\\
    
        \end{cases}
\end{equation}

\section{Dissimilarity ablations}
\label{apx:dissimilarity_ablations}
Additionally to the experiments in the main body, we provide additional information and ablations. The additional ablations include the effect of changing the dissimilarity weight $\lambda$ and metric, the number of layers used for regularization and the architectures. 

\subsection{Layers}
\begin{figure}
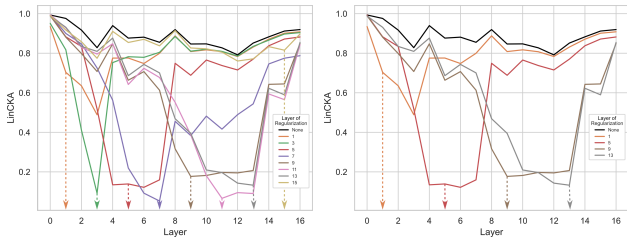

    \centering
    \begin{minipage}{0.49\linewidth}
         \centering
         \includegraphics[width=\linewidth]{figures/different_layers/diagonal_different_layer_all.pdf}
     \end{minipage}
     \hfill
     \begin{minipage}{0.49\linewidth}
         \centering
         \includegraphics[width=\linewidth]{figures/different_layers/diagonal_different_layers_selective.pdf}
  
     \end{minipage}
   
    \caption{\textit{LinCKA} between two ResNet34s with the last model being regularized on the former at a specific architecture position. Left: Diagonal \textit{LinCKA} values of all plots from  \cref{fig:diff_layers_modded_v2}. Right: Diagonal \textit{LinCKA} values of a smaller subset for better visibility.}
    \label{fig:ablation_position_in_architecture_diagonal}
\end{figure}

Complimentary to \cref{sec:dissimilarity_ablation}, we provide the diagonal \textit{LinCKA} similarity of the models from \cref{fig:diff_layers_modded_v2}. When choosing a layer for regularization, the \textit{LinCKA} similarity decreases the most at the selected layer, mostly affecting layers that are part of its residual block. After a downsampling stage, the similarity tends to increase drastically, but still stays below baseline levels (see \cref{fig:ablation_position_in_architecture_diagonal}).

\subsection{Regularizing multiple layers}
\begin{figure}
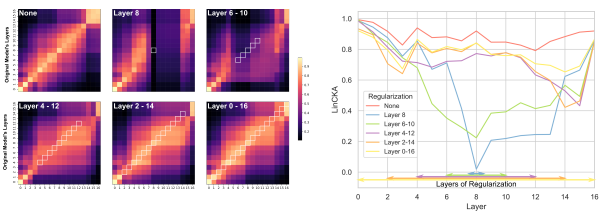

    \centering
    \begin{minipage}{0.49\linewidth}
         \centering
         \includegraphics[width=\linewidth]{figures/multilayer/multiple_layers.pdf}
         \label{fig:multiple_layers}
     \end{minipage}
     \hfill
     \begin{minipage}{0.49\linewidth}
         \centering
         \includegraphics[width=\linewidth]{figures/multilayer/multilayer_diagonal_modded.pdf}
         \label{fig:multiple_layers_diagonal}
     \end{minipage}

    \caption{Regularizing at multiple layers leads to decreases in a wider area but to a lower overall decrease than single selected layers.}
    \label{fig:multi_layer_regularization}
\end{figure}

Additionally to the previously described setting, one can choose to enforce dissimilarity at multiple layers simultaneously as opposed to only a single selected layer. When choosing multiple layers, the decrease in similarity from baseline is lower than when regularizing only a single layer, as the dissimilarity loss is averaged over all layers. This is highlighted in \cref{fig:multi_layer_regularization}. It nicely portraits the decrease of dissimilarity as more layers are regularized at once.

\subsection{ResNet101 and CIFAR100}
We expand the experiments to deeper architectures on CIFAR100, as previous experiments were constrained to CIFAR10 and ResNet34.

\begin{figure}
    \centering
    \includegraphics[width=.9\linewidth]{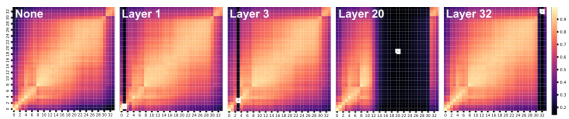}
    \caption{\textit{LinCKA} values of ResNet101s regularized at different layers by minimizing \textit{LinCKA} between models on CIFAR100.}
    \label{fig:resnet101_cifar100}
\end{figure}

\paragraph{Layer regularization}
Similarly to \cref{sec:dissimilarity_ablation} we regularize ResNet101 at a very early location Layer $1$, an early layer $3$, a intermediate Layer $20$, and a later layer $32$, visualized in \cref{fig:resnet101_cifar100}. Similarly to the smaller ResNet34 we observe that learned dissimilarity is either localized to the explicit layer of choice or affects the layers of the residual block the regularized layer is situated in. This is especially prominent for the 23 layer deep residual block when regularizing Layer $20$.

\subsection{Evaluating different dissimilarity metrics and loss weights}
In our experiments we so far constrained ourselves to \textit{ExpVar} and \textit{LinCKA} with $\lambda = 1$. In order to evaluate if this regularization weight is too strong or too weak, we ablate it for $\lambda \in \{0.25,1.0, 4.0\}$ in \cref{tab:exp_var} \cref{tab:LinCKA_weights}, and  \cref{tab:L2Corr}.
Across all metrics one can see that increasing $\lambda$  decreases Cohen's $\kappa$ and increases $JSD$ while simultaneously decreasing the accuracy of the new model. Setting $\lambda$ too high can lead to instabilities, and significantly reduced ensembling performance as is the case for \textit{L2Corr}. Furthermore we can observe that different layers require different $\lambda$ values as no one value dominates the others for the entire architecture, complicating the process, should one want to optimize $\lambda$.

\begin{table}[]
    \centering
    \caption{Output metrics of ResNet34 regularized through \textit{ExpVar} at various layers.}
    \resizebox{.95\linewidth}{!}{
    \begin{tabular}{lllrrrrr}
\toprule
Metric & Layer & Loss Weight &  Ensemble Acc &  New Model Acc  &  Cohens Kappa  &    JSD  &     ERD  \\
\midrule
ExpVar &     1 &        0.25 &       95.590 \textpm 0.154 &        94.866 \textpm 0.198 &       48.006 \textpm 1.237 & 2.887 \textpm 0.088 & 33.113 \textpm 3.534\% \\
ExpVar &     1 &        1.00 &       95.584 \textpm 0.191 &        94.404 \textpm 0.366 &       44.961 \textpm 2.389 & 3.412 \textpm 0.263 & 36.627 \textpm 5.932\% \\
ExpVar &     1 &        4.00 &       95.536 \textpm 0.091 &        93.716 \textpm 1.561 &       41.594 \textpm 7.017 & 4.113 \textpm 1.495 & 46.085 \textpm 11.372\% \\
ExpVar &     3 &        0.25 &       95.286 \textpm 0.180 &        94.036 \textpm 0.384 &       47.923 \textpm 3.376 & 3.187 \textpm 0.346 & 35.620 \textpm 5.063\% \\
ExpVar &     3 &        1.00 &       95.512 \textpm 0.149 &        94.194 \textpm 0.283 &       44.867 \textpm 1.541 & 3.389 \textpm 0.135 & 40.767 \textpm 5.076\% \\
ExpVar &     3 &        4.00 &       95.408 \textpm 0.166 &        94.244 \textpm 0.461 &       46.050 \textpm 1.636 & 3.304 \textpm 0.162 & 37.850 \textpm 2.064\% \\
ExpVar &     8 &        0.25 &       95.552 \textpm 0.126 &        94.838 \textpm 0.239 &       48.798 \textpm 1.321 & 2.823 \textpm 0.019 & 35.708 \textpm 2.977\% \\
ExpVar &     8 &        1.00 &       95.586 \textpm 0.230 &        94.336 \textpm 0.254 &       43.303 \textpm 1.521 & 3.584 \textpm 0.060 & 42.741 \textpm 3.601\% \\
ExpVar &     8 &        4.00 &       95.656 \textpm 0.188 &        94.240 \textpm 0.310 &       42.671 \textpm 0.769 & 3.756 \textpm 0.196 & 38.315 \textpm 2.720\% \\
ExpVar &    13 &        0.25 &       95.454 \textpm 0.139 &        94.698 \textpm 0.165 &       48.873 \textpm 1.717 & 2.865 \textpm 0.077 & 32.910 \textpm 0.984\% \\
ExpVar &    13 &        1.00 &       95.482 \textpm 0.280 &        93.574 \textpm 1.132 &       40.243 \textpm 4.897 & 4.519 \textpm 1.412 & 43.949 \textpm 7.067\% \\
ExpVar &    13 &        4.00 &       95.268 \textpm 0.125 &        93.742 \textpm 0.335 &       43.711 \textpm 0.620 & 3.747 \textpm 0.180 & 40.178 \textpm 3.466\% \\
\bottomrule
\end{tabular}
}
    \label{tab:exp_var}
\end{table}

\begin{table}[]
\centering
 \caption{Output metrics of ResNet34 regularized through \textit{LinCKA} at various layers.}
\resizebox{.95\linewidth}{!}{
\begin{tabular}{lllrrrrr}
\toprule
Metric & Layer & Loss Weight &  Ensemble Acc [\%] &  New Model Acc [\%] &   Cohens Kappa [\%] & JSD [\%] &     ERD [\%] \\
\midrule
LinCKA &     1 &        0.25 &       95.656 \textpm 0.143 &        94.660 \textpm 0.256 &       47.593 \textpm .933 & 3.170 \textpm 0.212 & 32.005 \textpm 2.659 \\
LinCKA &     1 &        1.00 &       95.538 \textpm 0.153 &        94.534 \textpm 0.146 &       46.511 \textpm .382 & 3.336 \textpm 0.391 & 32.972 \textpm 3.915 \\
LinCKA &     1 &        4.00 &       95.585 \textpm 0.093 &        94.450 \textpm 0.261 &       46.532 \textpm .352 & 3.336 \textpm 0.218 & 32.477 \textpm 5.028 \\
LinCKA &     3 &        0.25 &       95.536 \textpm 0.194 &        94.838 \textpm 0.302 &       51.064 \textpm .022 & 2.740 \textpm 0.084 & 33.170 \textpm 3.865 \\
LinCKA &     3 &        1.00 &       95.564 \textpm 0.059 &        94.832 \textpm 0.069 &       49.281 \textpm .796 & 2.858 \textpm 0.173 & 28.752 \textpm 3.523 \\
LinCKA &     3 &        4.00 &       95.590 \textpm 0.131 &        94.852 \textpm 0.084 &       48.930 \textpm .531 & 2.893 \textpm 0.057 & 35.783 \textpm 3.018 \\
LinCKA &     8 &        0.25 &       95.520 \textpm 0.172 &        94.916 \textpm 0.305 &       49.681 \textpm .818 & 2.840 \textpm 0.149 & 29.145 \textpm 4.743 \\
LinCKA &     8 &        1.00 &       95.654 \textpm 0.184 &        94.950 \textpm 0.465 &       49.399 \textpm .225 & 2.843 \textpm 0.203 & 33.563 \textpm 1.808 \\
LinCKA &     8 &        4.00 &       95.604 \textpm 0.201 &        94.816 \textpm 0.234 &       47.883 \textpm .204 & 2.982 \textpm 0.153 & 34.566 \textpm 4.494 \\
LinCKA &    13 &        0.25 &       95.512 \textpm 0.115 &        94.838 \textpm 0.223 &       51.030 \textpm .478 & 2.708 \textpm 0.124 & 29.956 \textpm 1.436 \\
LinCKA &    13 &        1.00 &       95.490 \textpm 0.148 &        94.676 \textpm 0.132 &       49.758 \textpm .719 & 2.832 \textpm 0.125 & 34.973 \textpm 2.744 \\
LinCKA &    13 &        4.00 &       95.560 \textpm 0.062 &        94.720 \textpm 0.190 &       49.935 \textpm .917 & 2.854 \textpm 0.136 & 33.426 \textpm 4.583 \\
\bottomrule
\end{tabular}
}
    \label{tab:LinCKA_weights}
\end{table}

\begin{table}[]
    \centering
    \caption{Output metrics of ResNet34 regularized through \textit{L2Corr} at various layers.}
\resizebox{.95\linewidth}{!}{
\begin{tabular}{lllrrrrrr}
\toprule
Metric & Layer & Loss Weight &  Ensemble Acc  [\%] &  New Model Acc  [\%] &  Cohens Kappa [\%] &     JSD [\%] & ERD [\%] \\
\midrule
L2Corr &     1 &        0.25 &       94.878 \textpm 0.236 &        84.932 \textpm 3.072 &       20.040 \textpm  5.378 & 12.277   \textpm   3.154 &  78.008 \textpm 12.239 \\
L2Corr &     1 &        1.00 &       95.280 \textpm 0.372 &        70.014 \textpm 24.841 &       20.176 \textpm 26.374 & 27.264   \textpm  24.040 & 125.746 \textpm 91.014 \\
L2Corr &     1 &        4.00 &       94.770 \textpm 0.138 &        36.706 \textpm 37.475 &        1.727 \textpm  9.740 & 51.226   \textpm  28.757 & 358.925 \textpm 221.863 \\
L2Corr &     3 &        0.25 &       95.580 \textpm 0.306 &        94.492 \textpm 0.811 &       46.383 \textpm  3.550 &  3.291   \textpm   0.675 &  35.765 \textpm 8.759 \\
L2Corr &     3 &        1.00 &       95.636 \textpm 0.131 &        94.660 \textpm 0.453 &       46.918 \textpm  1.769 &  3.205   \textpm   0.337 &  34.598 \textpm 5.420 \\
L2Corr &     3 &        4.00 &       95.346 \textpm 0.595 &        75.138 \textpm 37.168 &       31.375 \textpm 24.993 & 19.251   \textpm  29.950 & 133.428 \textpm 197.775 \\
L2Corr &     8 &        0.25 &       95.652 \textpm 0.088 &        94.920 \textpm 0.135 &       48.895 \textpm  2.401 &  2.877   \textpm   0.098 &  36.460 \textpm 3.253 \\
L2Corr &     8 &        1.00 &       95.638 \textpm 0.173 &        94.864 \textpm 0.192 &       47.767 \textpm  1.476 &  2.976   \textpm   0.119 &  33.190 \textpm 2.801 \\
L2Corr &     8 &        4.00 &       95.684 \textpm 0.142 &        94.854 \textpm 0.157 &       48.318 \textpm  2.705 &  3.015   \textpm   0.145 &  32.675 \textpm 4.037 \\
L2Corr &    13 &        0.25 &       95.532 \textpm 0.163 &        94.718 \textpm 0.175 &       50.558 \textpm  1.417 &  2.768   \textpm   0.080 &  33.224 \textpm 2.386 \\
L2Corr &    13 &        1.00 &       95.512 \textpm 0.155 &        94.684 \textpm 0.312 &       49.418 \textpm  0.510 &  2.829   \textpm   0.081 &  37.031 \textpm 3.395 \\
L2Corr &    13 &        4.00 &       95.322 \textpm 0.094 &        93.846 \textpm 0.281 &       47.663 \textpm  1.640 &  3.329   \textpm   0.180 &  39.130 \textpm 4.778 \\
\bottomrule
\end{tabular}
}
\label{tab:L2Corr}
\end{table}

\subsection{More detailed output metrics}
Revisiting \cref{eq:full_loss_untrained} the factor lambda determines the importance of learning representations that are dissimilar. Given that we observe 
\Cref{tab:ensemble_performance_main} in the main did not contain all metrics of interest we gathered, hence we provide the table containing additional metrics like the Jensen-Shannon-Divergence (JSD) (see \cref{apx:output_metrics}).

\begin{table}
\centering
\caption{Ensembles of ResNet34, regularized at various layers with the \textit{ExpVar} metric. Best metrics are displayed bold, while metrics better than the baseline are underlined.
\\}
\resizebox{.95\linewidth}{!}{
\begin{tabular}{ccccccc}
\toprule
 \multirow{2}{*}{\textbf{Output metric}} &  \multicolumn{2}{c}{\textbf{Dissimilarity}} & \multicolumn{4}{c}{\textbf{Ensemble of N Models}}\\
 \cmidrule{4-7}
 & \textbf{Metric} & \textbf{Layer} & 2 & 3 & 4 & 5 \\
\midrule
\multirow{5}{*}{\textbf{Ensemble Acc. [$\%$]} $\uparrow$} & \textbf{Baseline} & - & 95.438 \textpm 0.087 & 95.730 \textpm 0.082 & 95.880 \textpm 0.073 & 95.936 \textpm 0.081 \\
& \multirow{4}{*}{\textbf{ExpVar}} & 1 & \underline{95.580} \textpm 0.189 & \textbf{95.914} \textpm 0.039 & \underline{96.014} \textpm 0.048 & \textbf{96.130} \textpm 0.042 \\
&  & 3 & \underline{95.543} \textpm 0.147 & \underline{95.877} \textpm 0.118 & \underline{95.968} \textpm 0.112 & \underline{95.975} \textpm 0.038 \\
&  & 8 & \underline{95.570} \textpm 0.010 & \underline{95.897} \textpm 0.006 & \textbf{96.050} \textpm 0.092 & \underline{96.103} \textpm 0.025 \\
&  & 13 & \textbf{95.748} \textpm 0.114 & \underline{95.906} \textpm 0.113 & \underline{95.970} \textpm 0.029 & \underline{96.034} \textpm 0.076 \\
\midrule
\multirow{5}{*}{\textbf{$Acc_U$  [$\%$]} $\uparrow$} & \textbf{Baseline} & - & \textbf{94.708} \textpm 0.223 &  94.828 \textpm 0.165 & 94.918 \textpm 0.163 & \textbf{94.898} \textpm 0.134 \\
& \multirow{4}{*}{\textbf{ExpVar}} & 1 & 94.426 \textpm 0.292& 94.810 \textpm 0.079& \textbf{94.942} \textpm 0.100& \textbf{94.896} \textpm 0.302 \\
& & 3 & 94.325 \textpm 0.254 & \textbf{94.903} \textpm 0.043 & 94.875 \textpm 0.186 & 94.832 \textpm 0.163 \\
& & 8 & 94.403 \textpm 0.146 & 94.733 \textpm 0.377 & 94.707 \textpm 0.095 & 94.770 \textpm 0.425 \\
& & 13 & 94.608 \textpm 0.304 &  94.772 \textpm 0.173 & 94.828 \textpm 0.278 & 94.636 \textpm 0.204  \\

\midrule
\multirow{5}{*}{$\kappa$ [$\%$] $\downarrow$} & \textbf{Baseline} & - & 52.045 \textpm 0.707 & 51.852 \textpm 1.417 & 51.478 \textpm 1.226 & 51.588 \textpm 1.113 \\
& \multirow{4}{*}{\textbf{ExpVar}} & 1 & \underline{45.636} \textpm 1.696 & \underline{47.153} \textpm 1.162 & \underline{48.052} \textpm 0.683 & \underline{48.377} \textpm 1.035 \\
&  & 3  & \underline{45.472} \textpm 1.844 & \underline{46.818} \textpm 0.931 & \underline{48.125} \textpm 0.498 & \underline{48.601} \textpm 0.542 \\
&  & 8  & \underline{45.358} \textpm 0.839 & \textbf{46.438} \textpm \underline{0.801} & \textbf{46.838} \textpm 1.217 & \textbf{46.776} \textpm 1.590 \\
&  & 13  & \textbf{44.361} \textpm 0.803 & \underline{46.771} \textpm 0.708 & \underline{48.169} \textpm 0.772 & \underline{48.692} \textpm 0.327 \\

\midrule
\multirow{5}{*}{\textbf{JSD [$\%$] } $\uparrow$} & \textbf{Baseline} & - & 2.688 \textpm 0.037 & 2.698 \textpm 0.038 & 2.692 \textpm 0.014 & 2.672 \textpm 0.026 \\
& \multirow{4}{*}{\textbf{ExpVar}} & 1 & \underline{3.268} \textpm 0.113 &\underline{ 3.146} \textpm 0.153 & \textbf{3.030} \textpm 0.087 & \underline{2.992} \textpm 0.106 \\
&  & 3                                 & \underline{3.386} \textpm 0.227 & \underline{3.160} \textpm 0.157 & \textbf{3.036} \textpm 0.113 & \underline{2.978} \textpm 0.096 \\
&  & 8                                 & \textbf{3.501} \textpm 0.155 & \textbf{3.290} \textpm 0.183 & \underline{3.204} \textpm 0.207 & \textbf{3.164} \textpm 0.270\\
&  & 13                                & \textbf{3.500} \textpm 0.141 & \underline{3.253} \textpm 0.104 & \underline{3.115} \textpm 0.088 & \underline{3.034} \textpm 0.080 \\

\midrule
\multirow{5}{*}{\textbf{EDR [$\%$] } $\uparrow$} & \textbf{Baseline} & - & 31.596 \textpm  4.426 &  31.117 \textpm  2.883 &  30.907 \textpm  0.827 &  31.154 \textpm  1.536\\
& \multirow{4}{*}{\textbf{ExpVar}} & 1                                   & \textit{36.085} \textpm  4.440 &  \textit{34.867} \textpm  2.334 &  \textit{34.105} \textpm  0.374 &  \textit{33.454} \textpm  1.368\\
&  & 3                                                                   & \textit{37.314} \textpm  3.921 &  \textbf{36.548} \textpm  1.820 &  \textbf{34.816} \textpm  1.918 &  \textbf{34.583} \textpm  1.803\\
&  & 8                                                                   & \textbf{37.817} \textpm  1.288 &  35.971 \textpm  0.716 &  \textit{34.529} \textpm  0.765 &  \textit{34.157} \textpm  1.554\\
&  & 13                                                                  & \textit{34.944} \textpm  5.814 &  \textit{33.559} \textpm  2.560 &  \textit{33.154} \textpm  2.072 &  \textit{32.831} \textpm  1.691 \\

\bottomrule

\end{tabular}
}

\label{tab:ensemble_performance_apx}
\end{table}

\section{Choice of similarity metrics}
Many elaborate similarity metrics exist, from SVCCA \cite{Raghu2017} to PWCCA \cite{Morcos2018} to CKA \cite{Kornblith2019} a variety of methods can be employed to learn dissimilar representation as all are differentiable.

In this work we decided to keep the metric choice as simple as possible and use aligned regression and correlation metrics \textit{ExpVar} and \textit{L2Corr} as their channel-wise nature and alignment allows good interpretability and introspection possibilities for early development. 
Additionally we included \textit{LinCKA} as it is the most recent popular similarity metric and has a lower memory footprint than SVCCA, due to it working on the similarity matrices instead of representations directly. Additionally \textit{LinCKA} can be nicely calculated on mini-batches \cite{Nguyen2021}.

%

\end{document}